\documentclass[letterpaper]{article} 
\usepackage{aaai2026}
\usepackage{times}  
\usepackage{helvet}  
\usepackage{courier}  
\usepackage[hyphens]{url}  
\usepackage{graphicx} 
\usepackage{natbib}  
\usepackage{caption} 
\usepackage{algorithm}
\usepackage{algorithmic}
\usepackage{amsmath}
\usepackage{multirow}

\usepackage[table]{xcolor}
\usepackage{color}

\usepackage{array}
\usepackage{multirow}
\usepackage{booktabs}
\usepackage{pifont}
\usepackage{amssymb}
\usepackage{amsmath}
\usepackage{bm}
\definecolor{Gray}{gray}{0.93}
\usepackage{tikz}
\usepackage{newfloat}
\usepackage{listings}
\DeclareCaptionStyle{ruled}{labelfont=normalfont,labelsep=colon,strut=off} 
\floatstyle{ruled}
\newfloat{listing}{tb}{lst}{}
\floatname{listing}{Listing}
\title{Enhancing Supervised Composed Image Retrieval via Reasoning-Augmented Representation Engineering}
\author{
    Jun Li\equalcontrib, Hongjian Dou\equalcontrib, Zhenyu Zhang, Kai Li, Shaoguo Liu\thanks{Corresponding author}, Tingting Gao
}
\affiliations{
    \textsuperscript{} Kuaishou Technology\\

}

\usepackage{bibentry}

\begin{document}

\maketitle

\begin{abstract}
Composed Image Retrieval (CIR) presents a significant challenge as it requires jointly understanding a reference image and a modified textual instruction to find relevant target images.
Some existing methods attempt to use a two-stage approach to further refine retrieval results.
However, this often requires additional training of a ranking model.
Despite the success of Chain-of-Thought (CoT) techniques in reducing training costs for language models, their application in CIR tasks remains limited ---
compressing visual information into text or relying on elaborate prompt designs.
Besides, existing works only utilize it for zero-shot CIR, as it is challenging to achieve satisfactory results in supervised CIR with a well-trained model.
In this work, we proposed a framework that includes the \textbf{P}yramid  \textbf{M}atching Model with \textbf{T}raining-\textbf{F}ree \textbf{R}efinement (PMTFR) to address these challenges.
Through a simple but effective module called Pyramid Patcher, we enhanced the Pyramid Matching Model's understanding of visual information at different granularities.
Inspired by representation engineering, we extracted representations from COT data and injected them into the LVLMs.
This approach allowed us to obtain refined retrieval scores in the Training-Free Refinement paradigm without relying on explicit textual reasoning, further enhancing performance.
Extensive experiments on CIR benchmarks demonstrate that PMTFR surpasses state-of-the-art methods in supervised CIR tasks.
The code will be made public.
\end{abstract}


\section{Introduction}
In recent years, Composed Image Retrieval (CIR)~\cite{vo2019composing,goenka2022fashionvlp,wen2023target,bai2024sentencelevel} has gained increasing attention.
Compared to traditional image retrieval tasks~\cite{gordo2016deep,liu2016deepfashion,chen2022deep}, CIR requires not only a reference image as input but also a textual description of the expected modifications to the reference image.
The complexity of understanding multi-modal data and modeling cross-modal relationships presents technical challenges, but also enables novel applications of CIR methods in fields such as e-commerce~\cite {shankar2017deep} and \mbox{Internet search~\cite{deldjoo2023review}}.

\begin{figure}[t]
	\centering
	\includegraphics[width=1.0\linewidth]{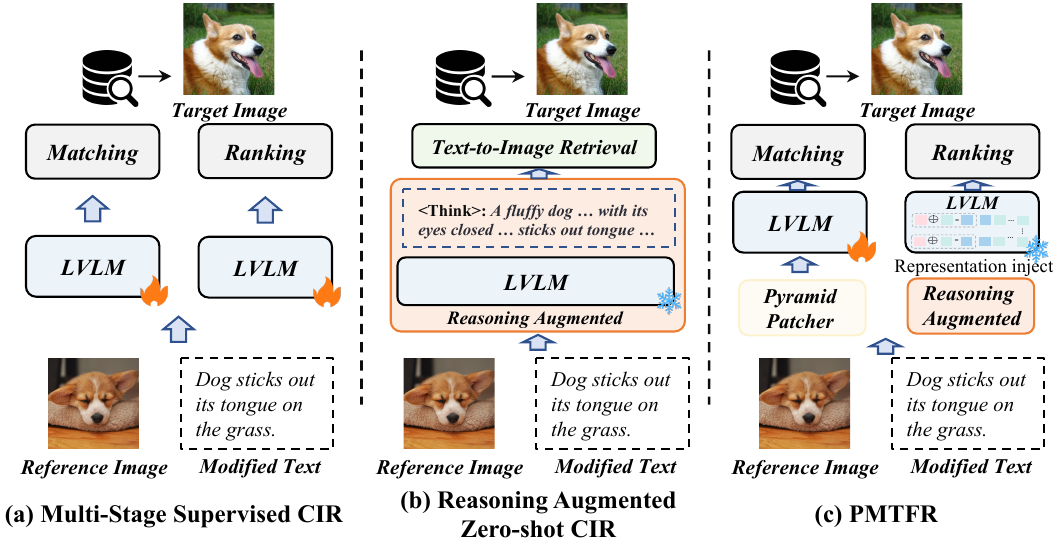}
         \caption{Comparison of Different Training Paradigms.
         (a) The Multi-stage supervised method requires additional training of a ranking model.
         (b) Utilizing the LVLMs to generate better reasoning text descriptions (Reasoning-Augmented) for zero-shot CIR tasks through a training-free approach.
         (c) The proposed PMTFR utilizes Pyramid Patcher to capture richer visual information and benefits the model in a training-free manner through the representation injection of the reasoning-augmented information.
         }
	\label{fig:intro}
\end{figure}
Large Vision-Language Models (LVLMs) have demonstrated promising multimodal understanding capabilities, and have been adopted in various multimodal tasks, including visual grounding~\cite{deng2021transvg}, visual question answering~\cite{antol2015vqa}, and CIR tasks~\cite{bai2024sentencelevel,sun2025leveraging,karthik2024vision,sun2023training}.
However, some of these works~\cite{karthik2024vision,tang2025reason} use LVLMs to extract better textual descriptions, which are then used to retrieve images, while others~\cite{sun2025leveraging,bai2024sentencelevel} rely on complex input prompt designs.
These cross-modal generation—such as image captioning—inevitably leads to information loss.
For that sake, these methods may fall short in CIR tasks which require both multi-granular visual information and cross-modal correlation understanding. 
For example, given the modified instruction `dog sticks out its tongue on the grass' (in Fig.~\ref{fig:intro}), the CIR models need to consider both macro-level background information `on the grass' and micro-level details of the dog's tongue.

Based on these considerations, we propose training a Pyramid Matching Model by utilizing a pre-trained LVLM.
By removing the large language model (LLM) head and taking the last token as representation, we simply input composed queries and target images, and use InfoNCE~\cite{he2020momentum} loss to minimize the distance between positive pairs while maximizing the distance between negative pairs.
This approach allows for better utilization of the world knowledge embedded in the pre-trained LVLM, while quickly adapting to the CIR task through straightforward alignment training.
To enhance the model's understanding of both fine-grained and coarse-grained visual information, we propose a module named Pyramid Patcher, inspired by the multi-scale technique in visual detection~\cite{singh2018analysis}.
Unlike the multi-scale technique, which randomly introduces images of different scales during training to adapt to varying object sizes during testing, Pyramid Patcher divides an image into multiple tokens with different visual receptive fields.
This approach significantly improves the capability of LVLM without introducing excessive computational overhead.
After training the Pyramid Matching Model, we can obtain an initial retrieval result for any composed query.

In supervised CIR tasks, the retrieval model is required to be trained on a training set within a specific domain and tested within it to evaluate performance.
Some existing multi-stage methods~\cite{liu2023candidate,liu2025lamra} additionally train a ranking model to reorder the retrieval results for more accurate results.
However, training a ranking model requires reconstructing the training data and consumes additional computational resources.
This leads us to consider \textit{whether it is possible to refine retrieval results without training}.
Achieving this without training while providing informational gain often implies a stronger generalization ability.
This directs our attention to Chain-of-Thought~\cite{wei2022chain}, which can enhance the capabilities of models without additional training.
In zero-shot CIR tasks~\cite{yang2024ldre}, some methods~\cite{sun2023training,tang2025reason} incorporate the Chain-of-Thought to generate more accurate textual descriptions of reference images and modified instructions.
After that, these methods utilized the CLIP~\cite{radford2021learning} model for text and image representation extraction to perform retrieval.
Based on these considerations, we designed a Training-Free Refinement paradigm, which adjusts the retrieval results of the Pyramid Matching Model, without the need to train an additional ranking model, thus \mbox{achieving further performance improvements.}

However, directly adopting such Reasoning methods presents two problems:
1) In zero-shot CIR tasks, these methods aim to produce better textual descriptions, while in our Training-Free Refinement stage, we aim to obtain a better refinement score.
2) Existing reasoning-based methods rely on explicit textual reasoning paths, which often lead to significant consumption of computing resources.
To address these empirical research gaps, we first pair the composed query with the candidate images retrieved by the Pyramid Matching Model, then utilize the pre-trained LVLM to determine whether they form a correct combination to obtain a refinement score.
Inspired by representation engineering~\cite{tang2025unlocking,zou2023representation}, which extracted the representations of LLMs with data that reflects specific capabilities, and treats these representations as the fundamental unit of analysis to understand and control high-level capabilities of LLMs.
We designed a reasoning-augmented representation (RAug-Rep) extraction and injection method in the Training-Free Refinement stage.
On the training set, we extract the `RAug-Rep' using pre-generated reasoning paths.
During the inference phase, we inject the `RAug-Rep' into the intermediate layers of the model to participate in the forward propagation of model.
Interestingly, this approach effectively improves the accuracy of the refinement scores, as if finding a key to unlock a certain capability of the model.
By simply inserting it into the model, we can stimulate a certain capability of the model, thereby achieving a performance improvement without inputting any explicit textual reasoning paths.
Our main contributions \mbox{are summarized as follows:}
\begin{itemize}
    \item We proposed a novel framework that includes the \textbf{P}yramid  \textbf{M}atching Model with \textbf{T}raining-\textbf{F}ree \textbf{R}efinement (PMTFR), which enhances the model's understanding of both coarse-grained and fine-grained visual information and employs a novel training-free paradigm to further refine the retrieval results.
    \item To the best of our knowledge, we are the first to apply representation engineering to the CIR task by extracting representations from reasoning paths and injecting them into the model, which provides an interesting direction for exploration within the CIR community.
    \item We propose the Pyramid Patcher module, which is simple yet quite effective.
    It provides the model with tokens of different visual receptive fields and \mbox{leads to performance improvements.}
    \item Extensive experiments are conducted on two commonly used benchmarks, which demonstrate that PMTFR outperforms all the state-of-the-art methods \mbox{in supervised CIR tasks.}
    
\end{itemize}

\section{Related Works}
\subsection{Composed Image Retrieval}
Composed Image Retrieval (CIR)~\cite{vo2019composing} is a challenging task due to its reliance on multimodal information for target image retrieval.
Most works focus on how to better integrate multimodal information, such as early-fusion~\cite{wen2024simple,levy2024data} and late-fusion~\cite{baldrati2022conditioned,chen2024spirit}.
Some methods~\cite{liu2023candidate,liu2025lamra} employ a multi-stage paradigm, where a ranking model is trained in the second stage to further enhance retrieval performance.
However, training a ranking model requires reconstructing the training data and consumes additional computational resources.
With the rapid development of Chain-of-Thought~\cite{wei2022chain}, many works~\cite{sun2023training,tang2025reason,yang2024ldre} utilize this technique in zero-shot CIR tasks to generate better text descriptions.
Nevertheless, these reasoning-augmented texts depend on complex prompt designs and explicit reasoning paths, posing significant challenges for implementation in embedding-based retrieval tasks like supervised CIR.
Based on this, we propose the PMTFR, which combines the advantages of multi-stage and reasoning-augment, achieving \mbox{better performance with higher efficiency.}

\subsection{Large Vision-Language Models}
Large Vision-Language Models (LVLMs) have made significant strides in multimodal learning, such as BLIP~\cite{li2022blip}, LLaVA~\cite{liu2023visual}, and Qwen-VL~\cite{bai2023qwen}.
The LVLMs have achieved excellent results in various multimodal tasks, including visual grounding~\cite{deng2021transvg} and visual question answering~\cite{antol2015vqa}, among others.
In recent years, some CIR methods~\cite{sun2025leveraging,bai2024sentencelevel,li2025encoder} have started leveraging the capabilities of LVLMs.
They use LVLMs to enhance model performance by generating better textual descriptions~\cite{tang2025reason}, sentence-level prompting~\cite{bai2024sentencelevel}, or soft prompting~\cite{sun2025leveraging}.
These works mostly focus on complex designs for text, thereby overlooking the exploration of rich visual cues.
In contrast, we leverage the LVLM as an encoder, simply inputting the composite query and target image to obtain the corresponding representations.
To enhance the understanding of visual information, we propose a straightforward Pyramid Patcher module, which enables the model to handle information at different visual granularities.

\subsection{Representation Engineering}
Representation Engineering (RepE) ~\cite{zou2023representation} extracted the representations of large language models (LLMs) with data that reflects specific capabilities, and treats these representations as the fundamental unit of analysis to understand and control high-level capabilities of LLMs.
As a widely applied technology, it is often used in LLMs and applied in many areas, such as hallucination alleviation~\cite{arditi2024refusal, li2023inference}, instruction following~\cite{stolfoimproving}, and reasoning~\cite{tang2025unlocking}.
In recent years, there have also been works~\cite{tian2025representation,mcgrath2022acquisition} beginning to explore its potential applications in LVLMs.
In this work, we pioneer the adaptation of RepE to composed image retrieval (CIR), where reasoning-augmented representations are extracted and injected via a training-free paradigm to \mbox{boost retrieval performance significantly.}

\section{Method}
\begin{figure*}[t]
	\centering
	\includegraphics[width=0.95\linewidth]{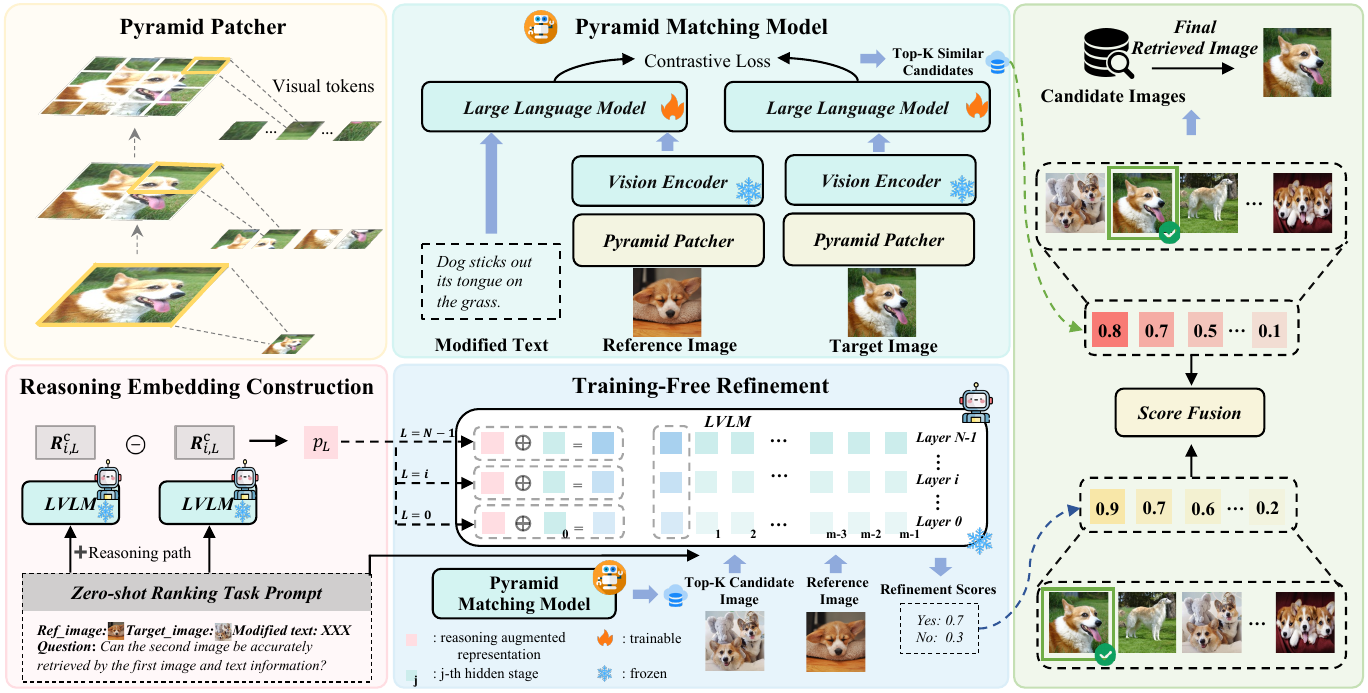}
         \caption{Overall pipeline of the proposed PMTFR.
         PMTFR is essentially a two-stage approach.
         In the first stage, the Pyramid Matching Model is used to obtain an initial retrieval result, where the Pyramid Patcher acquires tokens with different visual receptive fields for the visual input.
         The second stage is Training-Free Refinement, where representations are extracted from the reasoning data and then injected into the model. After that, we obtain more accurate correction scores, which are then fused with the scores from the initial retrieval result to serve as the basis for ranking.
         }
	\label{fig:framework}
\end{figure*}
\subsection{Preliminary}
\textbf{Task definition.}
Composed image retrieval (CIR) is a multimodal retrieval task that aims to retrieve the correct target image from a pool of candidate images based on a reference image and its corresponding modified text.
Specifically, the composed query is denote as $\mathcal{Q}_i = \left\{\boldsymbol{I}^r_i, \boldsymbol{T}_i\right\}$, where $\boldsymbol{I}^r_i$ denotes the reference image, and $\boldsymbol{T}_i$ denotes the modified text.
The goal of the CIR task is to retrieve the correct target image $\boldsymbol{I}_i^t$ from the candidate set $\mathcal{D}=$ $\left\{\boldsymbol{I}^t_0, \boldsymbol{I}^t_1, \ldots, \boldsymbol{I}^t_{C-1}\right\}$.

\subsection{PMTFR}
PMTFR consists of the Pyramid Matching Model and Training-Free Refinement as shown in Fig.~\ref{fig:framework}.
In the Pyramid Matching Model, the model learn a general representation of multimodal queries and images through a concise contrastive learning paradigm, which facilitates rapid retrieval.
To enhance the model's understanding of multi-granular visual information, we propose a module named Pyramid Patcher that provides the model with visual tokens of different receptive fields, substantially improving the capability of model.
Inspired by representation engineering, we extracted representations from Chain-of-Thought data and injected them into the Training-Free Refinement.
\\
\\
\textbf{Pyramid Patcher.}
In a standard Vision Transformer (ViT)~\cite{dosovitskiyimage}, an input 2D image $\boldsymbol{I} \in \mathbb{R}^{H \times W \times C}$ is divided into patches, resulting a sequence of flattened 2D patches $\boldsymbol{I}^p \in \mathbb{R}^{L \times\left(P^2 \cdot C\right)}$, where $\left(H, W\right)$ is the resolution of the original image, $C$ is the number of channels, $P$ is the patch size and $L = HW/P^2$.
The final token sequence fed into the visual encoder \mbox{can be formulated as:}
\begin{equation}
    \boldsymbol{I}^e \in \mathbb{R}^{L \times D} = \textrm{Emb}\left(\boldsymbol{I}^p\right),
\end{equation}
where $D$ denotes the dimension of the visual encoder, $\textrm{Emb}\left(\cdot\right)$ denotes the embedding layers.
In order to enhance the model’s understanding of multi-granular visual information, we propose the Pyramid Patcher that provides the model with visual tokens of different receptive fields.
To simultaneously obtain Tokens with $M$ different visual scales, we first duplicate the image $M$ times, i.e. $\mathcal{S} = \left\{\boldsymbol{I}_0, \boldsymbol{I}_1, \ldots, \boldsymbol{I}_M-1\right\}$.
We use a patch size of $P$ for the $\boldsymbol{I}_0$, $2 \times P$ for the $\boldsymbol{I}_1$ and $2^{M-1} \times P$ for the $\boldsymbol{I}_{M-1}$.
So the flattened 2D patches set can be formulated as:
\begin{equation}
    \mathcal{S}^p = \left\{ \boldsymbol{I}_i^p \in \mathbb{R}^{\frac{L}{2^{i}} \times\left(2^{i} \cdot P^2 \cdot C\right)} \middle| i \in \mathbb{Z}, \ 0 \leq i < M\right\},
    \label{eq:patcher}
\end{equation}
where $L$ is the number of visual tokens in the original image $\boldsymbol{I}_0$.
The final multi-scale visual tokens can be formulated as:
\begin{equation}
\begin{aligned}
    \boldsymbol{I}^e_{M} &= \textrm{cat}\left[\left\{\textrm{Emb}\left(\boldsymbol{I}_i^p\right) \middle| \boldsymbol{I}_i^p \in \mathcal{S}^p\right\}\right] \\
    &= \textrm{cat}\left[\left\{\boldsymbol{I}_{i}^e \in \mathbb{R}^{\frac{L}{2^{i}} \times D} \middle| i \in \mathbb{Z}, \ 0 \leq i < M\right\}\right],
\end{aligned}
\end{equation}
where the $\textrm{cat}[\cdot]$ denotes the concatenation operation performed along the first dimension.
After processing with Pyramid Patcher, for a visual token in $\boldsymbol{I}_{0}^e$ and $\boldsymbol{I}_{M-1}^e$, the former will contain more detailed visual information, while the latter will contain more macroscopic visual information.
\\
\\
\textbf{Pyramid Matching Model.}
The training objective of the model is to align the representations of the composed queries $\mathcal{Q} = \left\{\boldsymbol{I}^r, \boldsymbol{T}\right\}$ with the representations of the target images $\boldsymbol{I^t}$.
For the model $f_{\boldsymbol{\theta}}\left(\cdot\right)$, we take the hidden states of the last token as the representation, which \mbox{can be formulated as:}
\begin{equation}
\begin{aligned}
    \boldsymbol{u}^q &= f_{\boldsymbol{\theta}}\left(f_\zeta\left(\boldsymbol{I}_r\right),\boldsymbol{T}\right) \\
    \boldsymbol{u}^t &= f_{\boldsymbol{\theta}}\left(f_\zeta\left(\boldsymbol{I}_t\right)\right),
\end{aligned}
\end{equation}
where $f_\zeta\left(\cdot\right)$ denotes the Pyramid Patcher.
For a sample pair $\langle\boldsymbol{u}^q, \boldsymbol{u}^t\rangle$, we use the InfoNCE~\cite{he2020momentum} loss to minimize the similarity between the positive pairs and maximize between the negative pairs.
Specifically, let $N$ denote the size of the mini-batch and denote $\boldsymbol{\Psi}$ the set of all target image representations in a mini-batch, then the loss function can be expressed as:
\begin{equation}
   L = \sum_i^N-log\left( \frac{exp\left( \boldsymbol{u}^q_i \boldsymbol{u}^t_i /\tau\right)}
                          {\sum_{\boldsymbol{u}^t_j \in \boldsymbol{\Psi}} exp\left( \boldsymbol{u}^q_j, \boldsymbol{u}^t_j /\tau\right)  } \right),
\end{equation}
where $\tau$ is a temperature hyper-parameter.
During validation, for any composed query, we select the Top-N of the retrieval result according to the similarity score, where $s_i^m$ denotes the score of retrieval result $x_i$.
\begin{table*}[t]
\centering
\renewcommand\arraystretch{1.08}
\small
\begin{tabular}{l|c|cc|cc|cc|ccc}
\toprule[1pt]
\multirow{2}{*}{Method} & \multirow{2}{*}{Ref.} & \multicolumn{2}{c|}{Shirt} & \multicolumn{2}{c|}{Dress} & \multicolumn{2}{c|}{Tops\&Tees} & \multicolumn{3}{c}{Avg.}  \\ 
 && R@10 & R@50 & R@10 & R@50 & R@10 & R@50 & R@10 & R@50 & $R_{\text{mean}}$ \\
\hline
TIRG        &  CVPR'19  & 13.10 & 30.91 & 14.13 & 34.61 & 14.79 & 34.37 & 14.01 & 33.30 & 23.66 \\
ARTEMIS     &  ICLR'22  & 21.57 & 44.13 & 25.68 & 51.05 & 28.59 & 55.06 & 25.28 & 50.08 & 37.68 \\
TG-CIR      &  MM'23    & 52.60 & 72.52 & 45.22 & 69.66 & 56.14 & 77.10 & 51.32 & 73.09 & 62.16 \\
BLIP4CIR+Bi &  WACV'24  & 41.76 & 64.28 & 42.09 & 67.33 & 46.61 & 70.32 & 43.49 & 67.31 & 55.40 \\
Re-ranking  &  TMLR'24  & 50.15 & 71.25 & 48.14 & 71.34 & 55.23 & 76.80 & 51.17 & 51.17 & 62.15 \\
CASE        &  AAAI'24  & 48.48 & 70.23 & 47.44 & 69.36 & 50.18 & 72.24 & 48.79 & 70.68 & 59.74 \\
SPRC        &  ICLR'24  & 55.64 & 73.89 & 49.18 & 72.43 & 59.35 & 78.58 & 54.92 & 74.97 & 64.85 \\
CCIN        &  CVPR'25  & 55.93 & 74.14 & 49.38 & 72.58 & 57.93 & 77.56 & 54.41 & 74.76 & 64.59 \\
CIR-LVLM    &  AAAI'25  & \underline{58.59} & \underline{75.86} & 50.42 & 73.57 & 59.61 & 78.99 & 56.21 & 76.14 & 66.17 \\
ENCODER     &  AAAI'25  & 54.86 & 74.93 & \underline{51.51} & \textbf{76.95} & \textbf{62.01} & \textbf{80.88} & \underline{56.13} & \textbf{77.59} & \underline{66.86} \\
\rowcolor{Gray}PMTFR && \textbf{59.33} & \textbf{76.13} & \textbf{53.63} & \underline{75.47} & \underline{60.63} & \underline{80.32} & \textbf{57.86} & \underline{77.31} & \textbf{67.59}  \\ 
\bottomrule[1pt]
\end{tabular}
\caption{Comparison with the state-of-the-art methods on the validation set of Fashion-IQ.
where $R_{\text{mean}}$ indicates the average results across all the metrics.
The best results are in \textbf{boldface}, while the second-best results are \underline{underlined}.}
\label{tab:fashioniq_result}
\end{table*}
\\
\\
\noindent\textbf{Reasoning-Augmented Representation Construction.}
Recent works such as DeepSeek-R1~\cite{guo2025deepseek} have demonstrated the surprising generalization capabilities of Chain-of-Thought (COT)~\cite{wei2022chain} across many tasks.
However, this ability relies on explicit textual reasoning, which is not naturally compatible with embedding-based retrieval approaches such as CIR.
Fortunately, with the help of representation engineering~\cite{zou2023representation}, we extract an `RAug-Rep' that benefits from the reasoning ability of the model and enables model performance gains.
Specifically, for a positive pair $\langle \boldsymbol{Q}_i, \tilde{\boldsymbol{I}}_i^t \rangle$ in the training set, we first use the 'Refinement Prompt' (see Appendix A) to construct the question $\boldsymbol{q_i}$.
We then employ the 'Reasoning Path Construction Prompt' (see Appendix A) to generate input for GPT-4o~\cite{hurst2024gpt}, obtaining the reasoning path denoted as $\boldsymbol{c}_i$.
For each of these inputs, we feed them separately into the pre-trained LVLM to extract layer-wise representations.
Formally, for the standalone question $\boldsymbol{q_i}$ and the augmented input $(\boldsymbol{q_i};\boldsymbol{c}_i)$ (where $;$ denotes concatenation), the corresponding representations can be formulated as:
\begin{equation}
\label{equ:2 embeddings}
\boldsymbol{R}_{i,L}^q=h_L^{-1}\left(\boldsymbol{q}_i\right) \quad \boldsymbol{R}_{i,L}^c=h_L^{-1}\left(\boldsymbol{q}_i ; \boldsymbol{c}_i\right),
\end{equation}
where $h_L^{-1}\left(\boldsymbol s\right)$ denotes the hidden states corresponding to the last token in the token sequence $\boldsymbol s$ at the layer $L$ in LLM.
We take the $\boldsymbol{R}_{i,L}^c-\boldsymbol{R}_{i,L}^q$ as the `RAug-Rep' and average it over all samples in the training set, which \mbox{can be formulated as:}
\begin{equation}
\boldsymbol{p}_L=\frac{1}{| \boldsymbol\Phi|} \sum_{i=1}^{|\boldsymbol \Phi|}\left(\boldsymbol{R}_{i,L}^c-\boldsymbol{R}_{i,L}^q\right),
\end{equation}
where $\boldsymbol \Phi$ denotes all the samples in the training set.
\\
\\
\noindent\textbf{Training-Free Refinement.}
In the Training-Free Refinement, we inject `RAug-Rep' into a pre-trained LVLM to get the refinement score of the pair $\langle\boldsymbol{Q}_i, x_i\rangle$, where $\boldsymbol{Q}_i$ denotes the composed query in the validation set and $x_i$ denotes one of the Top-N images retrieved by the Pyramid Matching Model.
For the pair $\langle\boldsymbol{Q}_i, x_i\rangle$, we use the Refinement prompt (more details in Appendix) to form the question $\boldsymbol{q}_i'$, and the `RAug-Rep' will be injected into certain layers of the model.
We input $\boldsymbol{q}_i'$ to a pre-trained LVLM to get the probability (i.e., refinement score) of the [YES] token.
For more discussions on the selection of layers, please refer to Tab.~\ref{tab:inject pos}.
For a specific layer $L$ in the language model, we add the `RAug-Rep' of the corresponding layer to the first token of \mbox{the hidden states in that layer:}
\begin{equation}
\tilde{\boldsymbol{h}}_L^0=\boldsymbol{h}_L^0+ \alpha \cdot \boldsymbol{p}_L,
 \label{Eq:alpha}
\end{equation}
where $\boldsymbol{h}_L^0$ denotes the 0th token of the hidden states in layer $L$ and $\alpha$ is the hyperparameter to balance the strength of injection.
After that, following~\cite{liu2024context}, we scaled the magnitude of the vectors back to their original scale:
\begin{equation}
\hat{\boldsymbol{h}}_L^0=\tilde{\boldsymbol{h}}_L^0 \cdot \frac{\|\boldsymbol{h}_L^0\|_2}{\|\tilde{\boldsymbol{h}}_L^0\|_2}.
\end{equation}
We use the probability that the model outputs [YES] as the refinement score, denoted as $s_i^r$.
Finally, for each retrieval candidate, the final score can be formulated as:
\begin{equation}
    s_i = \lambda s_i^r + \left(1 - \lambda\right) s_i^m,
    \label{Eq:lambda}
\end{equation}
where $\lambda$ is the refinement strength.
We use the final scores $s_i$ as the basis for sorting the final retrieval results.

\label{sec:method}

\section{Experiments}

\subsection{Datasets and Metrics}
Following SPRC~\cite{bai2024sentencelevel}, we have selected the two most commonly used benchmarks for the supervised CIR task: Fashion-IQ~\cite{Wu_2021_fashioniq} and CIRR~\cite{liu2021CIRR}.
(1) \textbf{Fashion-IQ} comprises 77,684 fashion images, which are carefully organized into 30,134 triplets and categorized into three distinct subsets: Dress, Shirt, and Toptee.
Following previous work~\cite{bai2024sentencelevel,sun2025leveraging}, we trained our model on the training set and reported the Recall@K metrics for three subsets on the validation set.
(2) \textbf{CIRR} is a general image dataset that comprises 36,554 triplets.
Following previous work~\cite{bai2024sentencelevel,liu2021CIRR}, we trained our model in the training set and performed hyperparameter selection on the validation set.
Finally, we submitted the prediction results of the test set to an online server to obtain the Recall@K metrics \mbox{on the entire set and subsets.}

\subsection{Implementation Details}
We used the pre-trained Qwen2-VL-7B~\cite{bai2023qwen} model to initialize the weights of the Pyramid Matching Model and employed the AdamW optimizer, with a learning rate set to 4e-5 for 2 epochs.
The temperature in InfoNCE was set to 0.005, with a batch size of 1024 for the Fashion-IQ dataset and 512 for the CIRR dataset.
N is set to 100 in the selection of the top N candidates.
The pre-trained Qwen2-VL-7B is used in the Training-Free Refinement
All hyperparameters were selected using the validation set and held constant across experiments. 
All hyperparameter selections were made on the validation set and applied to all experiments.
All experiments were conducted on \mbox{a single NVIDIA A800 machine.}
\begin{table*}[t]
\centering
\renewcommand\arraystretch{1.08}
\small
\begin{tabular}{l|c|cccc|ccc|c}
\toprule[1pt]
\multirow{2}{*}{Method} & \multirow{2}{*}{Ref.}& \multicolumn{4}{c|}{Recall@K} & \multicolumn{3}{c|}{Recall$_{\text{subset}}$@K} & \multirow{2}{*}{Avg.} \\
&& K=1   & K=5   & K=10  & K=50  & K=1       & K=2       & K=3       &      \\ 
\hline
TIRG                    & CVPR'19 & 14.61 & 48.37 & 64.08 & 90.03 & -- & -- & -- & -- \\
ARTEMIS                 & ICLR'22 & 16.96 & 46.10 & 61.31 & 87.73 & 39.99 & 62.20 & 75.67 & 43.05 \\
TG-CIR                  & MM'23   & 45.25 & 78.29 & 87.16 & 97.30 & 72.84 & 89.25 & 95.13 & 75.57 \\
BLIP4CIR+Bi             & WACV'24 & 40.15 & 73.08 & 83.88 & 96.27 & 72.10 & 88.27 & 95.93 & 72.59 \\
Re-ranking              & TMLR'24 & 50.55 & 81.75 & 89.78 & 97.18 & 80.04 & 91.90 & 96.58 & 80.90 \\
CASE                    & AAAI'24 & 48.00 & 79.11 & 87.25 & 97.57 & 75.88 & 90.58 & 96.00 & 77.50 \\
SPRC                    & ICLR'24 & 51.96 & 82.12 & 89.74 & 97.69 & \textbf{80.65} & 92.31 & 96.60 & 81.38 \\
CCIN                    & CVPR'25 & 53.41 & \underline{84.05} & \textbf{91.17} & \underline{98.00} & -- & -- & -- & -- \\
ENCODER                 & AAAI'25 & 46.10 & 77.98 & 87.16 & 97.64 & 76.92 & 90.41 & 95.95 & 77.45 \\
CIR-LVLM                & AAAI'25 & \underline{53.64} & 83.76 & 90.60 & 97.93 & 79.12 & \underline{92.33} & \underline{96.67} & \underline{81.44} \\
\rowcolor{Gray}PMTFR     &         & \textbf{55.01} & \textbf{84.72} & \underline{91.06} & \textbf{98.58} & \underline{80.60} & \textbf{92.75} & \textbf{96.84} & \textbf{82.66} \\
\bottomrule[1pt]
\end{tabular}
\caption{Comparison with the state-of-the-art methods on the test set of CIRR, where Avg. indicates the average results of Recall@5 and Recall$_{\text{subset}}$@1. The best results are in \textbf{boldface}, while the second-best results are \underline{underlined}..}
\label{tab:cirr_result}
\end{table*}
\subsection{Experimental Results}
To evaluate the superiority of PMTFR, we compare our work with established baselines such as TIRG~\cite{vo2019tirg}, ARTEMIS~\cite{delmasartemis}, TG-CIR~\cite{wen2023target}, SPRC~\cite{bai2024sentencelevel},bi~\cite{liu2024bi}, CASE~\cite{levy2024data}, DWC~\cite{huang2024dynamic} and FashionERN~\cite{chen2024fashionern}, multi-stage work including Re-ranking~\cite{liu2023candidate}, as well as with the latest accepted works, including CIR-LVLM~\cite{sun2025leveraging}, Encoder~\cite{li2025encoder}, CCIN~\cite{tian2025ccin}.
\\
\\
\noindent\textbf{Comparisons to Prior Arts.}
As reported in Tab.~\ref{tab:fashioniq_result} and Tab.~\ref{tab:cirr_result}, PMTFR outperformed previous methods in terms of average recall metrics on both the Fashion-IQ and CIRR datasets.
Traditional methods based on ResNet, such as TIRG and ARTEMIS, performed significantly worse compared to many recent methods based on LVLMs.
Compared to CIR-LVLM, which also utilizes LVLMs, our approach outperformed it across all 9 metrics on the Fashion-IQ dataset, achieving an average improvement of \textbf{1.42\%}.
Similarly, on the CIRR dataset, our method surpassed CIR-LVLM in all 7 metrics, with an average gain of \textbf{1.22\%}.
In addition, unlike CIR-LVLM, which introduced task prompts and soft prompts that occupy excessive input context length, our method only utilized modified text and image information as input, which ensured higher training efficiency.
Notably, compared to the two-stage Re-ranking method, our method achieved an average improvement of \textbf{5.44\%} on the Fashion-IQ dataset and \textbf{1.76\%} on the CIRR dataset, without additional training of ranking model.
These results further demonstrate \mbox{the superiority of our proposed method.}
\subsection{Ablation Study \& Analysis}
\begin{figure}[t]
	\centering
	\includegraphics[width=1.0\linewidth]{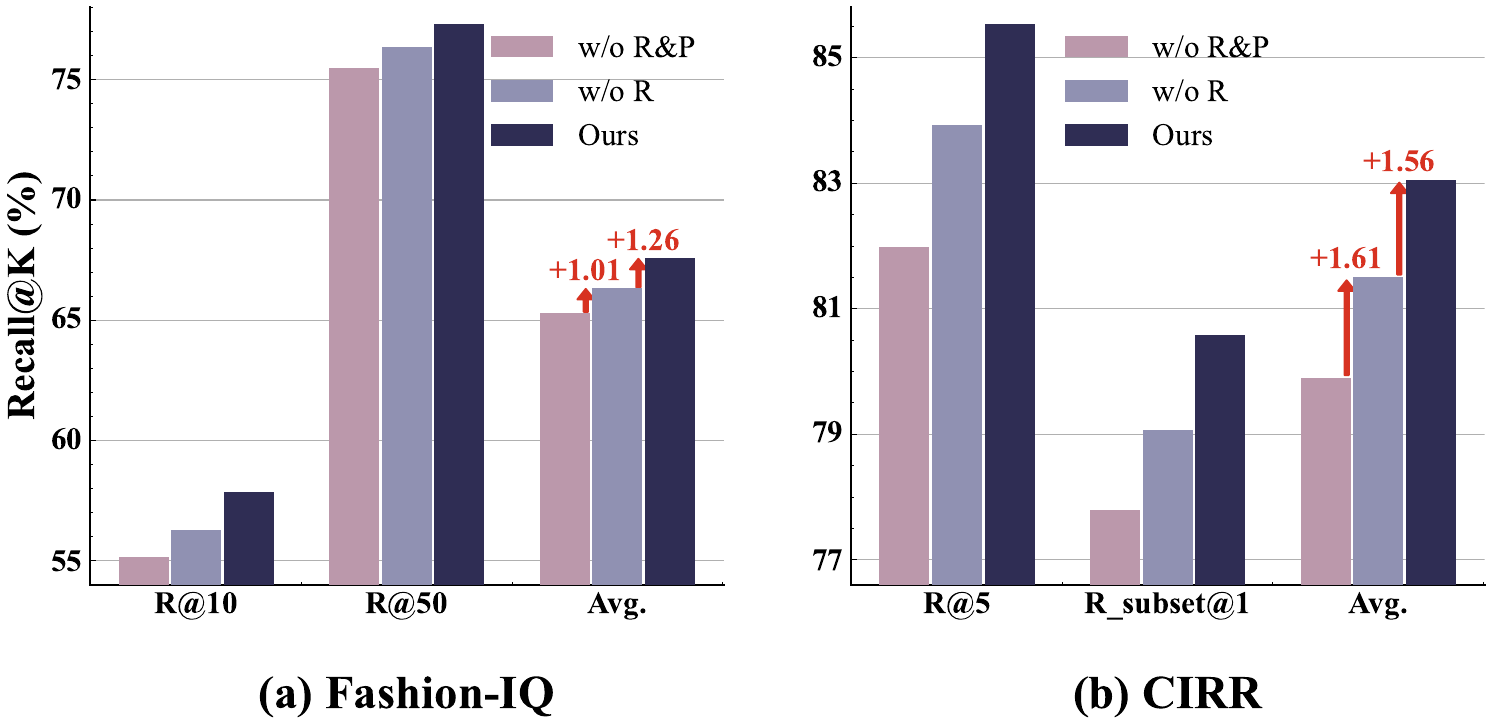}
         \caption{Ablation on Pyramid Patcher (denoted as P) and Training-Free Refinement (denoted as R).}
	\label{fig:ablation}
\end{figure}
\begin{table}[t]
\centering
\renewcommand\arraystretch{1.08}
\small
\begin{tabular}{l|ccc|c}
\toprule[1pt]
M & R@10  & R@50 & $R_{\text{mean}}$ & Token Length \\ 
\hline
1 & 55.15 & 75.49 & 65.32 & 165 \\
2 & 55.54 & 75.19 & 65.37 & 213 \\
3 & 55.35 & 75.86 & 65.60 & 225 \\
4 & 55.37 & 75.91 & 65.64 & 229 \\
5 & \textbf{56.30} & \textbf{76.36} & \textbf{66.33} & 231 \\
\bottomrule[1pt]
\end{tabular}
\caption{Comparison to different $M$ of Pyramid Pather in matching stage.}
\label{tab:patch level M}
\end{table}
\begin{figure}[t]
	\centering
	\includegraphics[width=1.0\linewidth]{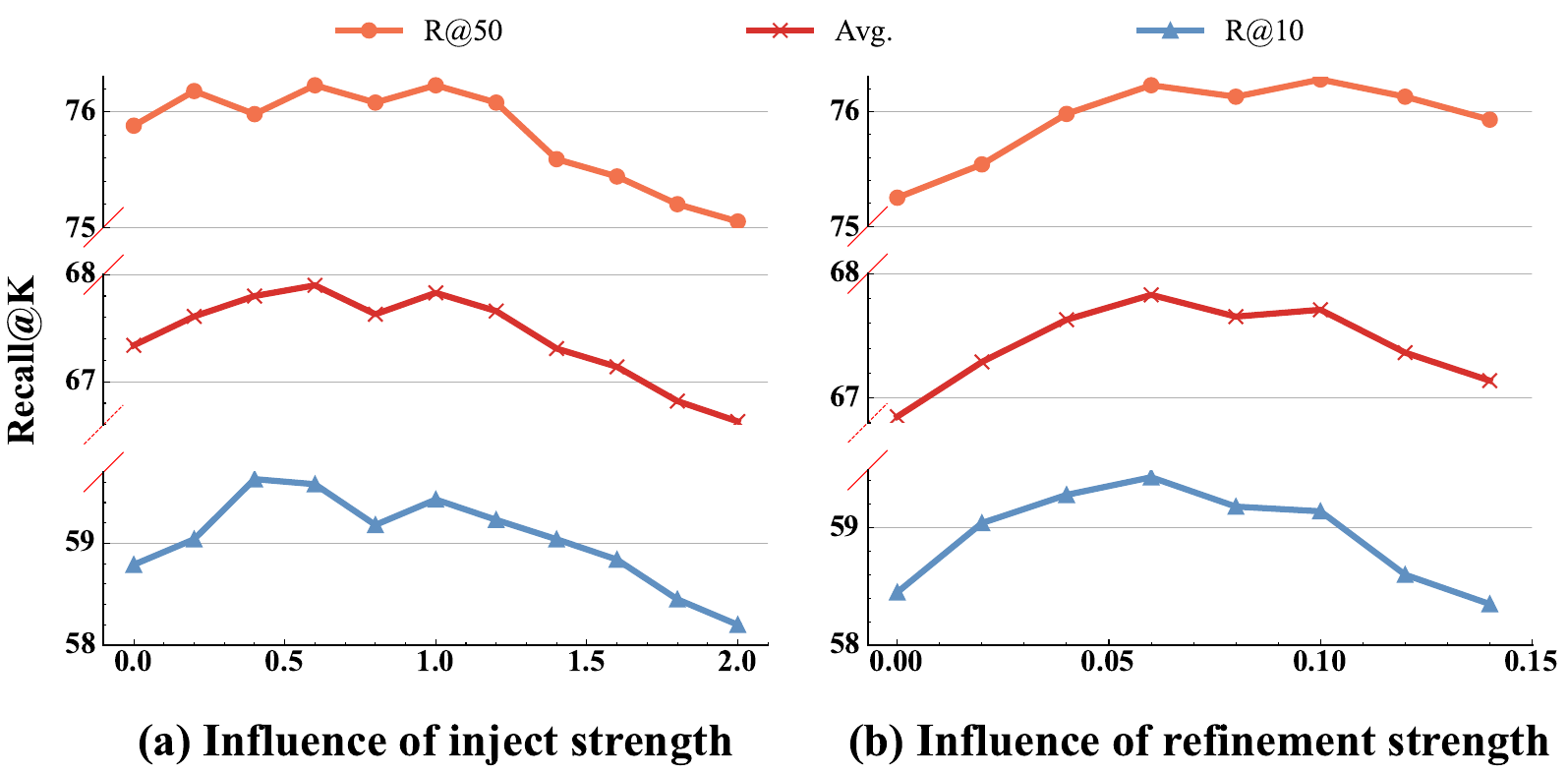}
         \caption{Sensitivity Analysis of different (a) inject strength $\alpha$ and (b) refinement strength $\lambda$ in Training-Free Refinement on the validation set of Fashion-IQ shirt.}
	\label{fig:inject_influence}
\end{figure}
\textbf{Effect of Pyramid Patcher and Training-Free Refinement.}
In Fig.~\ref{fig:ablation}, we conducted ablation experiments on Pyramid Patcher and Training-Free Refinement in the validation sets of Fashion-IQ and CIRR, and both parts provide notable performance gains.
Pyramid Patcher brings an average metric improvement of \textbf{1.01\%} in the Fashion-IQ dataset and \textbf{1.61\%} in the CIRR dataset.
Training-Free Refinement further delivers an average metric improvement of \textbf{1.26\%} in the Fashion-IQ and \textbf{1.56\%} in the CIRR.
A more detailed analysis of both parts will be presented later.
\\
\\
\noindent\textbf{Different visual scales of Pyramid Patcher.}
As shown in Eq.~\ref{eq:patcher}, the number of visual levels is denoted by $M$ in Pyramid Patcher.
We conducted experiments on Fashion-IQ using different $M$ and reported the corresponding recall metrics and number of image Tokens.
As shown in Tab.~\ref{tab:patch level M}, larger $M$ can obtain more diverse visual receptive field tokens and achieve better performance.
Since a larger patch size results in fewer image tokens, the number of image tokens does not increase significantly with the increase in $M$.
Compared to not using Pyramid Patcher ($M=1$), we achieved an improvement of \textbf{1.01\%} when $M=5$.
Finally, we chose the $M$ of 5 as the default setting.
\\
\\
\noindent\textbf{Sensitivity Analysis.}
We further analyze the contributions of Training-Free Refinement on Fashion-IQ shirt, examining two key hyperparameters: representation injection strength ($\alpha$) in Eq.~\ref {Eq:alpha}, and refinement score weighting ($\lambda$) in Eq.~\ref{Eq:lambda}.
As shown in Fig~\ref{fig:inject_influence}~(a), model performance improves significantly with moderate injection of `RAug-Rep'.
This indicates that these representations derived from explicit reasoning paths contain valuable information that activates latent model capabilities.
However, excessive injection strength disrupts the original input distribution, degrading performance due to representation misalignment.
The $\alpha=0.6$ brings the best performance, which is set to the default configuration for all experiments in this work.
Regarding the refinement strength, the best performance was achieved at 0.06 as shown in Fig~\ref{fig:inject_influence} (b).
Interestingly, even with a very small refinement strength (0.06) applied to adjust the matching scores, the average recall metric improved by \textbf{0.98\%}.
This demonstrates the effectiveness of the Training-Free Refinement.
\begin{table}[t]
\centering
\renewcommand\arraystretch{1.08}
\small
\begin{tabular}{c|ccc}
\toprule[1pt]
Inject Pos.& R@10  & R@50 & $R_{\text{mean}}$  \\ 
\hline
First & 58.64 & 75.49 & 67.07  \\
Middle & 58.89 & 75.69 & 67.29  \\
Last & 58.74 & 75.69 & 67.22  \\
All & 59.33 & 76.13 & 67.73  \\
\bottomrule[1pt]
\end{tabular}
\caption{Comparison to different inject positions of `RAug-Rep' in the Training-Free Refinement.}
\label{tab:inject pos}
\end{table}
\\
\\
\noindent\textbf{Different inject position in Training-Free Refinement.}
When extracting 'RAug-Rep', we capture layer-specific features to enable flexible injection position.
We empirically evaluated injection strategies on the Fashion-IQ shirt validation set using four configurations: \textbf{First} (Layer-0), \textbf{Middle} (Layer-14), \textbf{Last} (Layer-27), and \textbf{All} (Layer-0$\sim$27).
From the Tab.~\ref{tab:inject pos}, first-layer injection demonstrates a limited impact compared to middle/last-layer injection.
We hypothesize this occurs because the injected representation attenuates during forward diminishes through subsequent transformer layers.
Full-layer injection achieved optimal performance, indicating that single-layer modifications provide insufficient refinement.
Consequently, we adopt all-layer injection as the \mbox{default configuration across all experiments.}
\\
\\
\noindent\textbf{Visualization of representation.}
\begin{figure}[t]
	\centering
	\includegraphics[width=1.0\linewidth]{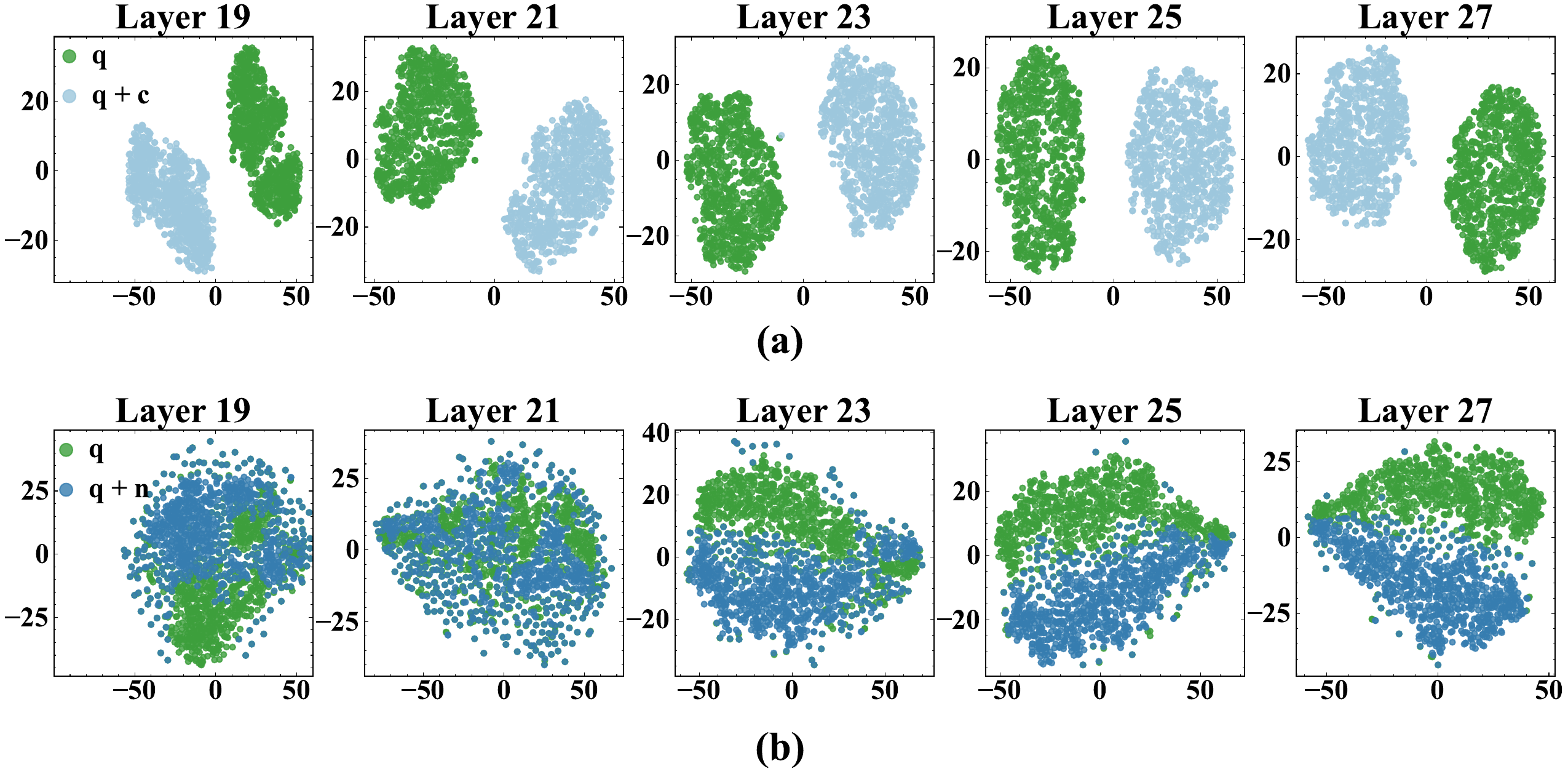}
         \caption{Visualization of the distribution of the last token from different layers using t-SNE. (a) inputting only the question (denoted as q) and the question along with the reasoning path (denoted as q + c); (b) inputting only the question (denoted as q) and the question along \mbox{with the noise text (denoted as q + n).}}
	\label{fig:emb_vis}
\end{figure}
\begin{table}[t]
\centering
\renewcommand\arraystretch{1.08}
\small
\begin{tabular}{c|ccc|cc}
\toprule[1pt]
Method & Training  & Inference & Tot. & R@10 & R@50\\ 
\hline
TR & 2.3h / epoch & 0.3h & 2.6h & 59.28 & 76.38\\
TFR & -- & 0.3h & 0.3h  & 59.33 & 76.13 \\
\bottomrule[1pt]
\end{tabular}
\caption{The time consumption and recall metrics of different training paradigms in the second stage. TR is short for Training Ranking model, TFR is short for \mbox{Training-Free Refinement.}}
\label{tab:time consum}
\end{table}
As shown in Fig.~\ref{fig:emb_vis}~(a), we visualized the $\boldsymbol{R}_{i,L}^q$ (q) and $\boldsymbol{R}_{i,L}^c$ (q + c) mentioned in Eq.~\ref{equ:2 embeddings} using the t-SNE~\cite{maaten2008visualizing} and selected the last several layers for display.
Meanwhile, we replaced the reasoning path with noisy text (q + n), and illustrated its distribution in Fig.~\ref{fig:emb_vis}~(b).
It can be observed that inputs containing richer informational content (reasoning paths) yield representations with greater distributional divergence from the original.
This indicates that the information density of inputs manifests in their representation distributions. 
Based on these, during the reasoning-augmented representation Construction phase, we used the $\boldsymbol{R}_{i,L}^c - \boldsymbol{R}_{i,L}^q$ as the final `RAug-Rep'.
The aim is to utilize this difference as a medium, allowing the model to acquire approximate capabilities without depending on explicit reasoning paths.
\\
\\
\noindent\textbf{Efficiency of the Training-Free Refinement.}
We conduct experiments on the Fashion-IQ shirt dataset to compare the efficiency and effectiveness of training a ranking model (denoted as TR) versus Training-Free Refinement (denoted as TFR).
While keeping the Matching Model unchanged, we constructed a dataset for ranking training using the top 10 recall results and trained a ranking model to sort these recall results.
In contrast, the Training-Free Refinement does not need any training.
As shown in Tab.~\ref{tab:time consum}, Training-Free Refinement can significantly reduce the time consumption while achieving results close to those of retraining the ranking model, demonstrating the effectiveness of the Training-Free Refinement.

\section{Conclusions and limitations}
In this work, we propose the Pyramid Matching Model with Training-Free Refinement (PMTFR).
By incorporating the Pyramid Patcher module, our model effectively captures both fine-grained and coarse-grained visual information, leading to performance improvement.
The Training-Free Refinement paradigm, which utilizes `RAug-Rep' extraction and injection, allows for refinement of retrieval results without the need for additional training.
While extensive experiments on widely-used benchmarks demonstrate that PMTFR outperforms existing state-of-the-art methods in supervised CIR tasks, there are still some limitations.
Although t-SNE visualization revealed that the representation distributions with and without Chain-of-Thought data form two clusters, there may be other reasonable methods worth exploring beyond simply subtracting the two representations.
The `RAug-Rep' is extracted and benefited from Chain-of-Thought data, and interestingly, they achieved performance improvements with generalization capabilities.
However, the specific mechanisms at play and their connection to the reasoning abilities of model are complex and were not deeply discussed in this work.
This remains a promising research direction that we hope will inspire future studies.

\bibliography{main}

@inproceedings{antol2015vqa,
 author = {Stanislaw Antol and
Aishwarya Agrawal and
Jiasen Lu and
Margaret Mitchell and
Dhruv Batra and
C. Lawrence Zitnick and
Devi Parikh},
 booktitle = {Proc. of ICCV},
 pages = {2425--2433},
 title = {{VQA:} Visual Question Answering},
 year = {2015}
}

@inproceedings{arditi2024refusal,
 author = {Andy Arditi and
Oscar Obeso and
Aaquib Syed and
Daniel Paleka and
Nina Panickssery and
Wes Gurnee and
Neel Nanda},
 booktitle = {Proc. of NeurIPS},
 title = {Refusal in Language Models Is Mediated by a Single Direction},
 year = {2024}
}

@article{bai2023qwen,
 author = {Bai, Jinze and Bai, Shuai and Yang, Shusheng and Wang, Shijie and Tan, Sinan and Wang, Peng and Lin, Junyang and Zhou, Chang and Zhou, Jingren},
 journal = {ArXiv preprint},
 title = {Qwen-vl: A frontier large vision-language model with versatile abilities},
 year = {2023}
}

@inproceedings{bai2024sentencelevel,
 author = {Yang Bai and
Xinxing Xu and
Yong Liu and
Salman Khan and
Fahad Khan and
Wangmeng Zuo and
Rick Siow Mong Goh and
Chun{-}Mei Feng},
 booktitle = {Proc. of ICLR},
 title = {Sentence-level Prompts Benefit Composed Image Retrieval},
 year = {2024}
}

@inproceedings{baldrati2022conditioned,
 author = {Alberto Baldrati and
Marco Bertini and
Tiberio Uricchio and
Alberto Del Bimbo},
 booktitle = {Proc. of CVPR},
 pages = {4955--4964},
 title = {Conditioned and composed image retrieval combining and partially fine-tuning
CLIP-based features},
 year = {2022}
}

@article{chen2022deep,
 author = {Chen, Wei and Liu, Yu and Wang, Weiping and Bakker, Erwin M and Georgiou, Theodoros and Fieguth, Paul and Liu, Li and Lew, Michael S},
 journal = {IEEE Transactions on Pattern Analysis and Machine Intelligence},
 pages = {7270--7292},
 title = {Deep learning for instance retrieval: A survey},
 year = {2022}
}

@inproceedings{chen2024fashionern,
 author = {Yanzhe Chen and
Huasong Zhong and
Xiangteng He and
Yuxin Peng and
Jiahuan Zhou and
Lele Cheng},
 booktitle = {Proc. of AAAI},
 pages = {1228--1236},
 title = {FashionERN: Enhance-and-Refine Network for Composed Fashion Image
Retrieval},
 year = {2024}
}

@article{chen2024spirit,
 author = {Chen, Yanzhe and Zhou, Jiahuan and Peng, Yuxin},
 journal = {ACM Transactions on Multimedia Computing, Communications and Applications},
 pages = {1--17},
 title = {Spirit: Style-guided patch interaction for fashion image retrieval with text feedback},
 year = {2024}
}

@article{deldjoo2023review,
 author = {Deldjoo, Yashar and Nazary, Fatemeh and Ramisa, Arnau and Mcauley, Julian and Pellegrini, Giovanni and Bellogin, Alejandro and Noia, Tommaso Di},
 journal = {ACM Computing Surveys},
 pages = {1--37},
 title = {A review of modern fashion recommender systems},
 year = {2023}
}

@inproceedings{delmasartemis,
 author = {Ginger Delmas and
Rafael Sampaio de Rezende and
Gabriela Csurka and
Diane Larlus},
 booktitle = {Proc. of ICLR},
 title = {{ARTEMIS:} Attention-based Retrieval with Text-Explicit Matching and
Implicit Similarity},
 year = {2022}
}

@inproceedings{deng2021transvg,
 author = {Jiajun Deng and
Zhengyuan Yang and
Tianlang Chen and
Wengang Zhou and
Houqiang Li},
 booktitle = {Proc. of ICCV},
 pages = {1749--1759},
 title = {TransVG: End-to-End Visual Grounding with Transformers},
 year = {2021}
}

@inproceedings{dosovitskiyimage,
 author = {Alexey Dosovitskiy and
Lucas Beyer and
Alexander Kolesnikov and
Dirk Weissenborn and
Xiaohua Zhai and
Thomas Unterthiner and
Mostafa Dehghani and
Matthias Minderer and
Georg Heigold and
Sylvain Gelly and
Jakob Uszkoreit and
Neil Houlsby},
 booktitle = {Proc. of ICLR},
 title = {An Image is Worth 16x16 Words: Transformers for Image Recognition
at Scale},
 year = {2021}
}

@inproceedings{goenka2022fashionvlp,
 author = {Sonam Goenka and
Zhaoheng Zheng and
Ayush Jaiswal and
Rakesh Chada and
Yue Wu and
Varsha Hedau and
Pradeep Natarajan},
 booktitle = {Proc. of CVPR},
 pages = {14085--14095},
 title = {FashionVLP: Vision Language Transformer for Fashion Retrieval with
Feedback},
 year = {2022}
}

@inproceedings{gordo2016deep,
 author = {Gordo, Albert and Almaz{\'a}n, Jon and Revaud, Jerome and Larlus, Diane},
 booktitle = {Proc. of ECCV},
 pages = {241--257},
 title = {Deep image retrieval: Learning global representations for image search},
 year = {2016}
}

@article{guo2025deepseek,
 author = {Guo, Daya and Yang, Dejian and Zhang, Haowei and Song, Junxiao and Zhang, Ruoyu and Xu, Runxin and Zhu, Qihao and Ma, Shirong and Wang, Peiyi and Bi, Xiao and others},
 journal = {ArXiv preprint},
 title = {Deepseek-r1: Incentivizing reasoning capability in llms via reinforcement learning},
 year = {2025}
}

@inproceedings{he2020momentum,
 author = {Kaiming He and
Haoqi Fan and
Yuxin Wu and
Saining Xie and
Ross B. Girshick},
 booktitle = {Proc. of CVPR},
 pages = {9726--9735},
 title = {Momentum Contrast for Unsupervised Visual Representation Learning},
 year = {2020}
}

@inproceedings{huang2024dynamic,
 author = {Fuxiang Huang and
Lei Zhang and
Xiaowei Fu and
Suqi Song},
 booktitle = {Proc. of AAAI},
 pages = {2303--2311},
 title = {Dynamic Weighted Combiner for Mixed-Modal Image Retrieval},
 year = {2024}
}

@article{hurst2024gpt,
 author = {Hurst, Aaron and Lerer, Adam and Goucher, Adam P and Perelman, Adam and Ramesh, Aditya and Clark, Aidan and Ostrow, AJ and Welihinda, Akila and Hayes, Alan and Radford, Alec and others},
 journal = {ArXiv preprint},
 title = {Gpt-4o system card},
 year = {2024}
}

@inproceedings{karthik2024vision,
 author = {Shyamgopal Karthik and
Karsten Roth and
Massimiliano Mancini and
Zeynep Akata},
 booktitle = {Proc. of ICLR},
 title = {Vision-by-Language for Training-Free Compositional Image Retrieval},
 year = {2024}
}

@inproceedings{levy2024data,
 author = {Matan Levy and
Rami Ben{-}Ari and
Nir Darshan and
Dani Lischinski},
 booktitle = {Proc. of AAAI},
 pages = {2991--2999},
 title = {Data Roaming and Quality Assessment for Composed Image Retrieval},
 year = {2024}
}

@inproceedings{li2022blip,
 author = {Junnan Li and
Dongxu Li and
Caiming Xiong and
Steven C. H. Hoi},
 booktitle = {Proc. of ICML},
 pages = {12888--12900},
 title = {{BLIP:} Bootstrapping Language-Image Pre-training for Unified Vision-Language
Understanding and Generation},
 year = {2022}
}

@inproceedings{li2023inference,
 author = {Kenneth Li and
Oam Patel and
Fernanda B. Vi{\'{e}}gas and
Hanspeter Pfister and
Martin Wattenberg},
 booktitle = {Proc. of NeurIPS},
 title = {Inference-Time Intervention: Eliciting Truthful Answers from a Language
Model},
 year = {2023}
}

@inproceedings{li2025encoder,
 author = {Li, Zixu and Chen, Zhiwei and Wen, Haokun and Fu, Zhiheng and Hu, Yupeng and Guan, Weili},
 booktitle = {Proc. of AAAI},
 pages = {5101--5109},
 title = {Encoder: Entity mining and modification relation binding for composed image retrieval},
 year = {2025}
}

@inproceedings{liu2016deepfashion,
 author = {Ziwei Liu and
Ping Luo and
Shi Qiu and
Xiaogang Wang and
Xiaoou Tang},
 booktitle = {Proc. of CVPR},
 pages = {1096--1104},
 title = {DeepFashion: Powering Robust Clothes Recognition and Retrieval with
Rich Annotations},
 year = {2016}
}

@inproceedings{liu2021CIRR,
 author = {Zheyuan Liu and
Cristian Rodriguez Opazo and
Damien Teney and
Stephen Gould},
 booktitle = {Proc. of ICCV},
 pages = {2105--2114},
 title = {Image Retrieval on Real-life Images with Pre-trained Vision-and-Language
Models},
 year = {2021}
}

@article{liu2023candidate,
 author = {Liu, Zheyuan and Sun, Weixuan and Teney, Damien and Gould, Stephen},
 journal = {ArXiv preprint},
 title = {Candidate set re-ranking for composed image retrieval with dual multi-modal encoder},
 year = {2023}
}

@inproceedings{liu2023visual,
 author = {Haotian Liu and
Chunyuan Li and
Qingyang Wu and
Yong Jae Lee},
 booktitle = {Proc. of NeurIPS},
 title = {Visual Instruction Tuning},
 year = {2023}
}

@inproceedings{liu2024bi,
 author = {Liu, Zheyuan and Sun, Weixuan and Hong, Yicong and Teney, Damien and Gould, Stephen},
 booktitle = {Proc. of WCACV},
 pages = {5753--5762},
 title = {Bi-directional training for composed image retrieval via text prompt learning},
 year = {2024}
}

@inproceedings{liu2024context,
 author = {Sheng Liu and
Haotian Ye and
Lei Xing and
James Y. Zou},
 booktitle = {Proc. of ICML},
 title = {In-context Vectors: Making In Context Learning More Effective and
Controllable Through Latent Space Steering},
 year = {2024}
}

@inproceedings{liu2025lamra,
 author = {Liu, Yikun and Zhang, Yajie and Cai, Jiayin and Jiang, Xiaolong and Hu, Yao and Yao, Jiangchao and Wang, Yanfeng and Xie, Weidi},
 booktitle = {Proc. of CVPR},
 pages = {4015--4025},
 title = {Lamra: Large multimodal model as your advanced retrieval assistant},
 year = {2025}
}

@article{maaten2008visualizing,
 author = {Maaten, Laurens van der and Hinton, Geoffrey},
 journal = {Journal of machine learning research},
 pages = {2579--2605},
 title = {Visualizing data using t-SNE},
 year = {2008}
}

@article{mcgrath2022acquisition,
 author = {McGrath, Thomas and Kapishnikov, Andrei and Toma{\v{s}}ev, Nenad and Pearce, Adam and Wattenberg, Martin and Hassabis, Demis and Kim, Been and Paquet, Ulrich and Kramnik, Vladimir},
 journal = {Proceedings of the National Academy of Sciences},
 pages = {e2206625119},
 title = {Acquisition of chess knowledge in AlphaZero},
 year = {2022}
}

@inproceedings{radford2021learning,
 author = {Alec Radford and
Jong Wook Kim and
Chris Hallacy and
Aditya Ramesh and
Gabriel Goh and
Sandhini Agarwal and
Girish Sastry and
Amanda Askell and
Pamela Mishkin and
Jack Clark and
Gretchen Krueger and
Ilya Sutskever},
 booktitle = {Proc. of ICML},
 pages = {8748--8763},
 title = {Learning Transferable Visual Models From Natural Language Supervision},
 year = {2021}
}

@article{shankar2017deep,
 author = {Shankar, Devashish and Narumanchi, Sujay and Ananya, HA and Kompalli, Pramod and Chaudhury, Krishnendu},
 journal = {ArXiv preprint},
 title = {Deep learning based large scale visual recommendation and search for E-Commerce},
 year = {2017}
}

@inproceedings{singh2018analysis,
 author = {Bharat Singh and
Larry S. Davis},
 booktitle = {Proc. of CVPR},
 pages = {3578--3587},
 title = {An Analysis of Scale Invariance in Object Detection {\-} {SNIP}},
 year = {2018}
}

@inproceedings{stolfoimproving,
 author = {Stolfo, Alessandro and Balachandran, Vidhisha and Yousefi, Safoora and Horvitz, Eric and Nushi, Besmira},
 booktitle = {Proc. of ICLR},
 title = {Improving Instruction-Following in Language Models through Activation Steering},
 year = {2024}
}

@article{sun2023training,
 author = {Sun, Shitong and Ye, Fanghua and Gong, Shaogang},
 journal = {ArXiv preprint},
 title = {Training-free zero-shot composed image retrieval with local concept reranking},
 year = {2023}
}

@inproceedings{sun2025leveraging,
 author = {Sun, Zelong and Jing, Dong and Yang, Guoxing and Fei, Nanyi and Lu, Zhiwu},
 booktitle = {Proc. of AAAI},
 pages = {7149--7157},
 title = {Leveraging large vision-language model as user intent-aware encoder for composed image retrieval},
 year = {2025}
}

@inproceedings{tang2025reason,
 author = {Tang, Yuanmin and Zhang, Jue and Qin, Xiaoting and Yu, Jing and Gou, Gaopeng and Xiong, Gang and Lin, Qingwei and Rajmohan, Saravan and Zhang, Dongmei and Wu, Qi},
 booktitle = {Proc. of CVPR},
 pages = {14400--14410},
 title = {Reason-before-retrieve: One-stage reflective chain-of-thoughts for training-free zero-shot composed image retrieval},
 year = {2025}
}

@inproceedings{tang2025unlocking,
 author = {Tang, Xinyu and Wang, Xiaolei and Lv, Zhihao and Min, Yingqian and Zhao, Wayne Xin and Hu, Binbin and Liu, Ziqi and Zhang, Zhiqiang},
 booktitle = {Proc. of ACL},
 title = {Unlocking General Long Chain-of-Thought Reasoning Capabilities of Large Language Models via Representation Engineering},
 year = {2025}
}

@inproceedings{tian2025ccin,
 author = {Tian, Likai and Zhao, Jian and Hu, Zechao and Yang, Zhengwei and Li, Hao and Jin, Lei and Wang, Zheng and Li, Xuelong},
 booktitle = {Proc. of CVPR},
 pages = {3974--3983},
 title = {CCIN: Compositional Conflict Identification and Neutralization for Composed Image Retrieval},
 year = {2025}
}

@article{tian2025representation,
 author = {Tian, Bowei and Lyu, Xuntao and Liu, Meng and Wang, Hongyi and Li, Ang},
 journal = {ArXiv preprint},
 title = {Why Representation Engineering Works: A Theoretical and Empirical Study in Vision-Language Models},
 year = {2025}
}

@inproceedings{vo2019composing,
 author = {Nam Vo and
Lu Jiang and
Chen Sun and
Kevin Murphy and
Li{-}Jia Li and
Li Fei{-}Fei and
James Hays},
 booktitle = {Proc. of CVPR},
 pages = {6439--6448},
 title = {Composing Text and Image for Image Retrieval - an Empirical Odyssey},
 year = {2019}
}

@inproceedings{vo2019tirg,
 author = {Nam Vo and
Lu Jiang and
Chen Sun and
Kevin Murphy and
Li{-}Jia Li and
Li Fei{-}Fei and
James Hays},
 booktitle = {Proc. of CVPR},
 pages = {6439--6448},
 title = {Composing Text and Image for Image Retrieval - an Empirical Odyssey},
 year = {2019}
}

@inproceedings{wei2022chain,
 author = {Jason Wei and
Xuezhi Wang and
Dale Schuurmans and
Maarten Bosma and
Brian Ichter and
Fei Xia and
Ed H. Chi and
Quoc V. Le and
Denny Zhou},
 booktitle = {Proc. of NeurIPS},
 title = {Chain-of-Thought Prompting Elicits Reasoning in Large Language Models},
 year = {2022}
}

@inproceedings{wen2023target,
 author = {Wen, Haokun and Zhang, Xian and Song, Xuemeng and Wei, Yinwei and Nie, Liqiang},
 booktitle = {Proc. of ACM MM},
 pages = {915--923},
 title = {Target-guided composed image retrieval},
 year = {2023}
}

@inproceedings{wen2024simple,
 author = {Haokun Wen and
Xuemeng Song and
Xiaolin Chen and
Yinwei Wei and
Liqiang Nie and
Tat{-}Seng Chua},
 booktitle = {Proc. of SIGIR},
 pages = {229--239},
 title = {Simple but Effective Raw-Data Level Multimodal Fusion for Composed
Image Retrieval},
 year = {2024}
}

@inproceedings{Wu_2021_fashioniq,
 author = {Hui Wu and
Yupeng Gao and
Xiaoxiao Guo and
Ziad Al{-}Halah and
Steven Rennie and
Kristen Grauman and
Rog{\'{e}}rio Feris},
 booktitle = {Proc. of CVPR},
 pages = {11307--11317},
 title = {Fashion {IQ:} {A} New Dataset Towards Retrieving Images by Natural
Language Feedback},
 year = {2021}
}

@inproceedings{yang2024ldre,
 author = {Zhenyu Yang and
Dizhan Xue and
Shengsheng Qian and
Weiming Dong and
Changsheng Xu},
 booktitle = {Proc. of SIGIR},
 pages = {80--90},
 title = {{LDRE:} LLM-based Divergent Reasoning and Ensemble for Zero-Shot Composed
Image Retrieval},
 year = {2024}
}

@article{zou2023representation,
 author = {Zou, Andy and Phan, Long and Chen, Sarah and Campbell, James and Guo, Phillip and Ren, Richard and Pan, Alexander and Yin, Xuwang and Mazeika, Mantas and Dombrowski, Ann-Kathrin and others},
 journal = {ArXiv preprint},
 title = {Representation engineering: A top-down approach to ai transparency},
 year = {2023}
}
\end{document}